\documentclass[10pt,twocolumn,letterpaper]{article}

\usepackage[pagenumbers]{wacv} 

\usepackage{graphicx}
\usepackage{amsmath}
\usepackage{amssymb}
\usepackage{booktabs}
\usepackage{mathtools}
\usepackage[table]{xcolor}
\usepackage{tikz}
\usetikzlibrary{patterns}

\definecolor{lightgray}{gray}{0.94}
\definecolor{darkgray}{gray}{0.9}

\DeclareMathOperator*{\argmax}{arg\,max}

\usepackage[pagebackref,breaklinks,colorlinks]{hyperref}

\usepackage[capitalize]{cleveref}
\crefname{section}{Sec.}{Secs.}
\Crefname{section}{Section}{Sections}
\Crefname{table}{Table}{Tables}
\crefname{table}{Tab.}{Tabs.}

\def\wacvPaperID{1926} 
\def\confName{WACV}
\def\confYear{2026}

\begin{document}

\title{Configurable Fairness: Direct Optimization of Parity Metrics via Vision-Language Models}

\author{Miao Zhang\\
New York University\\
{\tt\small miaozhng@nyu.edu}
\and
Rumi Chunara\\
New York University\\
{\tt\small rumi.chunara@nyu.edu}
}
\maketitle

\begin{abstract}
   Performance disparities of image recognition across demographic groups are known to exist in deep learning-based models, due to imbalanced group representations or spurious correlation between group and target labels. Previous work has addressed such challenges without relying on expensive group labels, typically by upweighting high-loss samples or balancing discovered clusters. However, these heuristic strategies lack direct connection to specific fairness metrics and cannot guarantee optimization of parity-based criteria like equal opportunity, which ensures equal chance to receive positive outcomes across groups. In this work, we propose a novel paradigm that directly optimizes parity-based fairness metrics through specifically designed training objectives, without requiring group labels. We leverage vision-language models to analyze sensitive attribute relevancy for individual samples, then formulate loss functions that mathematically connect to each target fairness metric. This enables flexible optimization of different fairness criteria based on application needs. Experiments on multiple image classification datasets show that our metric-specific approach significantly improves parity-based fairness criteria and outperforms existing methods.
\end{abstract}

\section{Introduction}
\label{sec:intro}

The ethical examination of machine learning systems is gaining increasing importance based on findings from real-world deployment of models. One aspect of this challenge is that models fit well specific to training data distributions, and predictions are shown to be biased towards majority samples~\cite{torralba2011unbiased}. Since in reality, data collection with a wide range of demographic representations is resource intensive and very difficult to be free of any biases, it is of great interest to develop model training algorithms to address fairness.

Early efforts in fair visual recognition have balanced model outcomes based on the group assignment of samples~\cite{gong2021mitigating,jung2023re, romano2020achieving, wang2020towards}. Given that obtaining sensitive attribute labels is difficult, approaches without using these labels have been developed, which typically include a bias discovery step, for example, identifying under-performed sample clusters based on model's latent representations, prediction errors, etc. By reweighting data or balancing training loss across identified clusters, the under-performed samples can be improved~\cite{Ahn2023mitigating, chakraborty2024exmap, chai2022fairness, espinosa2024efficient, lahoti2020fairness, liu2021just, seo2022unsupervised, wu2023discover, zhang2024discover}.

However, these approaches rely on heuristic strategies, including upweighting high-loss samples, balancing discovered clusters, or other proxy objectives. The strategies are effective but lack direct mathematical connection to specific fairness metrics. This disconnect is problematic because real-world applications often require specific notions of fairness. For example, equalized odds~\cite{hardt2016equality} ensures that different groups have equal chances of being predicted as each target class, crucial in risk assessments~\cite{berk2021fairness}, and equal opportunity ensures equal true positive rates across groups, essential in hiring contexts. Without directly optimizing these metrics, we cannot guarantee achieving the specific fairness criteria required by each application. To address this gap, we propose a paradigm shift: from proxy-based optimization to direct, metric-specific fairness optimization through mathematically grounded training objectives.

Building upon the success of large VLMs in mapping images and text to a shared latent space~\cite{jia2021scaling, radford2021learning}, recent research has increasingly explored their rich semantic knowledge. Pre-trained VLMs effectively capture the relationships between visual attributes and natural language descriptions, providing guidance for traditional vision tasks, such as classification~\cite{abdelfattah2023cdul, petryk2022guiding}, segmentation~\cite{he2023clip, wang2022cris}, retrieval~\cite{liu2021image}, and feature transformation~\cite{dunlap2023using, lou2022tecm}. Building on this well-established capability, we leverage VLMs to generate soft relevancy scores for visual attributes such as perceived \textit{race}, \textit{age}, and \textit{gender}, not as definitive classifications, but as regularization signals. This enables us to mathematically connect the optimization objective to target fairness metrics through attribute relevancy scores.

In this work, we establish a new paradigm for fairness-aware machine learning: fairness optimization should be metric-specific and direct rather than relying on proxy objectives. We focus on equal opportunity~\cite{hardt2016equality}, equalized odds~\cite{hardt2016equality}, and parity accuracy~\cite{berk2021fairness}, key parity-based metrics widely applied in real-world tasks~\cite{10.1145/3616865}. We propose a simple yet effective pipeline that directly optimizes chosen fairness metrics through specifically designed loss functions. Using the VLM-generated relevancy scores described above, we regularize target task learning by penalizing the correlation between class-wise model outputs and attribute scores, with penalty structures mathematically derived from each fairness metric's definition. Importantly, our approach remains effective even with imperfect and noisy VLM attribute signals. Our key contributions are as follows:

\begin{itemize}
\item We propose a novel framework that directly optimizes parity-based fairness metrics through mathematically grounded objectives, using penalty terms that encourage independence between model predictions and VLM-analyzed attribute scores.
\item Experiments on three datasets validate that our  method outperforms existing methods, achieving superior targeted fairness metrics while maintaining or improving model overall accuracy. 
\item We demonstrate robust fairness optimization even with imperfect attribute signals, showing that our regularization framework maintains effectiveness when VLM scores include errors or noises as in realistic settings.
\end{itemize}

\section{Related work}

\subsection{Fairness in Image Recognition} 
\noindent Algorithmic fairness considers inequity between sensitive groups in terms of aspects like model recognition accuracy, spurious label association~\cite{goyal2022fairness}, or robustness to noise and perturbation~\cite{dooley2022robustness, nanda2021fairness}.  
Most fairness approaches assume that sensitive attribute labels are available from the whole dataset. Further, this work addresses disentangling target and sensitive attribute information from image representations~\cite{park2021learning, park2022fair}, generalizing to worst cases~\cite{sagawa2019distributionally}, and data balance by resampling~\cite{agarwal2022does, li2019repair, yan2020fair} or augmentation~\cite{d2024improving, li2022cat, ramaswamy2021fair}.
However, the requirement on sample-level sensitive attribute labels are often unavailable in real-world data; due to legal or privacy concerns. Therefore, \textbf{fairness approaches unaware of sensitive attributes} are gaining more popularity, which is one of the motivations for our proposed method. In such existing approaches, researchers detect samples difficult to classify and influential to overall training costs in the model latent space and up-weight under-represented samples or clusters~\cite{Ahn2023mitigating, chai2022fairness, espinosa2024efficient, krishnakumar2021udis, seo2022unsupervised}. Alternatively, attribute classifiers are built by splitting samples into binary groups with maximized predicted probabilities~\cite{li2022discover}. These methods have shown efficacy on multiple image datasets through tackling worst-case samples to promote minority group performance. Further approaches identify biased attributes interpretable to humans by using synthesized images of generative models trained on the dataset~\cite{zhang2024distributionally}, or captions of mispredicted images~\cite{kim2024discovering}.   
We contribute to the set of approaches that do not require sample-level attribute labels beyond those provided for the target. Moreover, we propose robust and widely applicable objectives that directly enhance parity-based fairness, which has been relatively underexplored in the literature.

\subsection{Leveraging Vision-Language Models} 
\noindent \textbf{Interpretability, out-of-domain treatment, and bias discovery}. A few language-driven approaches implemented with VLMs have been explored for responsible computer vision tasks. For interpretability, researchers identify the influential attributes of high-error predictions using a descriptive prompt ~\cite{zhang2023drml, wang2023exploring}. For treating out-of-domain samples, text is used to drive the CLIP model to transfer images with feature varieties~\cite{shen2023clip}. For bias discovery, bias-aligning samples are located with CLIP embeddings as a prior~\cite{yenamandra2023facts}, then clustering is performed in the model latent space with high-dimensional embeddings. Other work generates captions for mis-predicted images~\cite{kim2023bias} or the whole dataset~\cite{zhang2024common} with VLMs, thus identifying bias from text. However, the setting highly depends on caption generation to extract potential responsible attributes. In general, studies that leverage VLMs for bias mitigation focus on any image features causing high-errors in a dataset; that goal is essentially different from our work which is to improve prediction parity among sample groups split by certain sensitive attributes.

\noindent \textbf{Label efficient learning}. Knowledge in VLMs has been leveraged in training settings of scarce data and labels, via natural language which is generally more available, or easy to compose. Language can guide image embedding transformations to align with unseen domains~\cite{zhang2022tip, dunlap2023using}. Researchers also guide model attention by language specification to image local regions, which enables task-related feature extraction in unsupervised training~\cite{zhong2022regionclip, petryk2022guiding, abdelfattah2023cdul}. In this work, we harness the power of VLMs for label-efficient learning in fairness applications. Specifically, we use VLM to align image samples with language descriptions of sensitive attributes, whose labels can be expensive to obtain. These alignment scores serve as effective regularization signals to promote sensitive attribute-independent predictions.

\section{Fairness Problem Formulation}
\label{sec:problem}

\noindent To formulate the fairness problem, we define an image dataset $\mathbf{X}\coloneqq\{\mathbf{x}^{(i)}\}_{i=1}^N$ with $N$ samples, a target $Y$, and a classifier $f$. $\mathbf{X}$ can be partitioned into sensitive groups based on the combination of the value of $Y$ and an attribute $A$ which is well-recognized for fairness discussion, \textit{e.g.}, demographic attributes. There are multiple parity-based fairness metrics that can be considered. In this work we focus on  
three commonly used metrics: equal opportunity, equalized odds, and accuracy parity. These metrics are relevant because they allow groups to differ in their base rates (e.g., the rates of positive or negative outcomes) while constraining disparities in error rates, which are generally considered more meaningful and useful for fair decision-making~\cite{hardt2016equality}. First, for a target class $Y=y$, we define the true positive rate gap across groups following~\cite{jung2022learning, shen2022optimising}: 
\begin{equation}\label{eq2}
\begin{aligned}
\small
TPR_y = \max_{\forall a_i, a_j \in A}|P (f(\mathbf{x}) = y |Y = y, A = a_i) \\
- P (f(\mathbf{x}) = y |Y = y, A = a_j)|.
\end{aligned}
\end{equation}
Equal opportunity metric measures the gap of positive (``advantaged'') outcome rate across groups~\cite{hardt2016equality} with respect to the positive label, denoted as:
\begin{equation} \label{eq3}
\small
    EOp = TPR_{1}.
\end{equation}
Equalized odds metric measures the gap in true prediction rate for both positive and negative class of a target~\cite{hardt2016equality}, which can be extended to multi-target setting~\cite{shen2022optimising, zhang2024distributionally} by:
\begin{equation} \label{eq4}
\small
    EOd = \sqrt{\frac{1}{|Y|}\sum_{y\in Y} TPR_y^2}.
\end{equation}
Lastly, accuracy parity measures the gap of the overall prediction accuracy (the rate of correct predictions) across groups~\cite{berk2021fairness}:
\begin{equation}
\small
    AP = \max_{\forall a_i, a_j \in A}|P (f(\mathbf{x}) = Y |A = a_i) - P (f(\mathbf{x}) = Y |A = a_j)|.
\end{equation}

\subsection{Why Configurable Fairness Matters}
\label{sec:motivation}


The fairness literature has established fundamental incompatibility between different parity-based metrics when base rates differ across groups~\cite{garg2020fairness, mashiat2022trade}. We demonstrate this through an example to motivate configurable fairness optimization.
Consider a binary classification scenario with two demographic groups: Group A (60\% of population, 50\% positive rate) and Group B (40\% of population, 25\% positive rate). Suppose a classifier yields: 

For Group A, $TPR=0.8$ (24/30 correct positives), $TNR=0.83$ (25/30 correct negatives). 

For Group B, $TPR=0.7$ (7/10 correct positives), $TNR=0.93$ (28/30 correct negatives) for Group B. 

In this case, the equal opportunity gap $EOp = 0.1$, and the accuracy parity gap $AP=0.058$. The incompatibility emerges when attempting to simultaneously satisfy the metrics: To achieve equal opportunity by raising Group B's $TPR$ to 0.8 increases its accuracy to 0.9, widening accuracy parity gap $AP$ to 0.083. Conversely, reducing Group A's $TPR$ to 0.7 lowers its accuracy to 0.767, increasing the gap to 0.108. This incompatibility reflects a mathematical constraint: since accuracy equals $TPR \times P(Y=1|A=a) + TNR \times P(Y=0|A=a)$ and groups have different positive rates $P(Y=1|A=a)$, forcing equal $TPR$ across groups while maintaining equal accuracy requires their $TNR$ to differ precisely to compensate, which rarely occurs naturally and makes satisfying multiple fairness metrics simultaneously impossible.

This incompatibility motivates our configurable framework: A one-size-fits-all approach might misalign with actual fairness requirements, so rather than treat fairness as a fixed constraint, we enable practitioners to optimize the specific metric most relevant to their applications.


\section{Method}
\label{sec:method}

\subsection{Preliminaries}
\label{sec:pre}
\noindent  Vision language models like CLIP~\cite{radford2021learning} are contrastively trained with image-text pairs. Model encoders learn to transform the image and the text description to the same embedding space for comparison. Specifically, cosine similarity $c$ is used to measure the correspondence between an image embedding $\vec{m}$ and a text embedding $\vec{d}$, by dividing their inner product by the product of $\ell2$ norms:
\begin{equation} \label{eq_cosine}
\small
c = \frac{\vec{m} \cdot \vec{d} }{ \lVert\vec{m}\rVert \lVert\vec{d}\rVert}.
\end{equation}

By leveraging $c$ to quantify how well an image matches a text description, the model has been used in image recognition tasks with zero-shot inference. We also experiment with other VLMs with similar function, BLIP~\cite{li2023blip} and ALIGN~\cite{jia2021scaling}, in a later section. We do not consider VLMs that focus on text generation instead of providing a direct image-text matching score, since we utilize $c$ to quantify sensitive attribute relevancy and build training objectives.

\textbf{Sensitive attribute estimation.} Assuming the type of sensitive attributes in question is pre-defined, we leverage CLIP retrieved correspondence $c$ between attribute description and image as a proxy of the sensitive information contained in an image. The handcrafted description set is denoted as $D = [d_i, i\in K]$, where $K$ is the number of attribute categories, $d_i$ is a sentence following the template used by the original CLIP model with an attribute embedded. In this work, we use $D = \{$\texttt{A photo of a young person}, \texttt{A photo of an old person}''$\}$ for \textit{age}, $D = \{$\texttt{A photo of a man}, \texttt{A photo of a woman}$\}$ for \textit{gender}, and $D = \{$\texttt{A photo of a white person, A photo of a person of color}$\}$ for \textit{race}. Additionally, we consider a non-human attribute \textit{color} of dogs or cats to validate method generalizability, with $D = \{$\texttt{A photo of a dog/cat with bright color, A photo of a dog/cat with dark color}$\}$. The specific categories for each attribute are selected corresponding to the sensitive groups studied in relevant literature, i.e., the attribute classes that the target model should be fair across. To quantify sample relevancy to an attribute, we compute the cosine similarity (equation~\ref{eq_cosine}) between the image and each text in the description set: $[c_i, i\in K]$. For attribute with binary categories ($K=2$), we use Softmax to compute score $\bar{c} = \frac{e^{c_1}}{e^{c1}+e^{c2}}$  following~\cite{wang2023exploring}: a larger $\bar{c}$ indicates closer match to the first category of the attribute and a smaller $\bar{c}$ indicates closer match to the second category. For $K>2$, we record the similarity to each category: $\bar{c_i} = \frac{e^{ci}}{e^{c1}+...+e^{c_K}}$ to compose training objectives as discussed in Section~\ref{sec:methodEop}.  

A natural concern is the reliability of the VLM-predicted score, which is continuous, in representing the discrete attribute partitions defined by domain experts. To address this, we do \textit{not} classify CLIP scores into predefined groups, as these groupings are often arbitrary~\cite{lingenfelter2022quantitative} (\textit{e.g.} the label \textit{young} in CelebA dataset is labeled based on subjective perception~\cite{liu2015deep}). Instead, we design fairness optimization objectives directly using the raw scores. This is motivated by the idea that a model whose predictions are independent of sample-level attribute relevancy will be fair across any group partition for that attribute. We formalize this intuition using an assumption from the capability of vision-language foundation models. We assume that training with large-scale and diverse data annotated by domain experts, foundation models are generalizable to perceive and reason semantics at human level for common scenes or objects~\cite{bommasani2021opportunities}, such as the sensitive attributes studied in this work: 

\textbf{Assumption 1} (Generalizability and perceptual robustness). Given an image $x$, $g_a(x)$ is how an attribute $a$ is perceived in the image. The pre-trained VLM image and text encoder is a good approximation to $g$: $(f_{img}, f_{text})\approx g$. 

Under this assumption, we aim to learn a target model whose prediction is independent of similarity score $\bar{c}$, which approximates the perception $g$ of an attribute in an image. This thus ensures that the model is independent of sensitive group assignment $h(g_a(x))$, regardless of the criteria $h$ used for group annotation. Next, we construct specific training objectives to optimize different fairness metrics.
 

\subsection{Optimizing Equal Opportunity}
\label{sec:methodEop}
\noindent We propose to maximize equal opportunity by minimizing the undesired correlation between predicted class probability and sensitive attribute relevancy, within the subset of positive class samples. The following loss function is used:
\begin{equation} \label{eop}
\mathcal{L}_{EOp} = \mathcal{L}_{ce} + \alpha|PCC (s_{Y=1}(\mathbf{x}), \bar{c} | Y=1)|,
\end{equation}
here $\mathcal{L}_{ce}$ denotes the average cross-entropy loss based on samples in the batch, $PCC$ is the Pearson correlation coefficient~\cite{cohen2009pearson}, $\alpha$ is a hyperparameter that controls the strength of the fairness regularization. The probability output of the target model's softmax function for class $y$, denoted as $s_{Y=y}(\mathbf{x})$, is used instead of the discrete class prediction $f(\mathbf{x})$, as the latter is non-differentiable and cannot be directly optimized for independence from $\bar{c}$. Here, $\bar{c}$ represents the vision-language model's predicted relevancy to the sensitive attribute. For binary attributes, $\bar{c}$ is a single value, and its correlation with $s_{Y=1}(\mathbf{x})$ is computed as $PCC (s_{Y=1}(\mathbf{x}), \bar{c})$, as described in Section~\ref{sec:pre}. For multi-category attributes, $\bar{c}$ consists of relevancy scores for each category, and we compute the mean correlation between $s_{Y=y}(\mathbf{x})$ and these scores.

By minimizing $\mathcal{L}_{EOp}$, ideally, $|PCC (s_{Y=1}(\mathbf{x}), \bar{c} | Y=1)| = 0$ can satisfy: 
\begin{align*}
\small
P (s_{Y=1}(\mathbf{x}) |Y = 1, a_i) 
= P (s_{Y=1}(\mathbf{x}) |Y = 1, a_j)
\end{align*}
Since predicted class label $f(\mathbf{x})$ is computed from the probability output, \textit{e.g.}, $f(\mathbf{x}) = \argmax(s_{Y=1}(\mathbf{x}))$ for classification, the condition above can satisfy: 
\begin{align*}
\small
P (f(\mathbf{x})=1 |Y = 1, a_i) 
= P (f(\mathbf{x})=1 |Y = 1, a_j),
\end{align*}
which ensures equal opportunity (equation~\ref{eq2},~\ref{eq3}).

\subsection{Optimizing Equalized Odds}

\noindent Equalized odds is stricter than equal opportunity, requiring equalized outcomes across groups for both positive and negative classes. The proposed loss function is: 
\begin{equation} \label{eod}
\mathcal{L}_{EOd} = \mathcal{L}_{ce} + \alpha\sum_{y \in Y}|PCC (s_{Y=y}(\mathbf{x}), \bar{c} | Y=y)|.
\end{equation}
The proof for $\mathcal{L}_{EOd}$ is similar to that for equal opportunity. The difference is that $\mathcal{L}_{EOd}$ extends the correlation coefficent penalty term to all target classes. Minimizing  $\mathcal{L}_{EOd}$ encourages the independence between positive prediction within each class and sensitive groups, directly optimizing for equalized odds criterion (equation~\ref{eq4}). 

\subsection{Optimizing Accuracy Parity}
\noindent Accuracy parity requires that the overall prediction accuracy is equal across different groups, regardless of class labels. We propose the loss function as:


\begin{equation} \label{ap}
\begin{aligned}
\mathcal{L}_{AP} = \mathcal{L}_{ce} + \alpha(\sum_{y \in Y}|PCC (s_{Y=y}(\mathbf{x}), \bar{c} | Y=y)| \\
+ \max_{\forall y_i, y_j \in Y}|Acc(Y=y_i) - Acc(Y=y_j)|),
\end{aligned}
\end{equation}
where $Acc$ denotes the prediction accuracy.
We show how $\mathcal{L}_{AP}$ is related to accuracy parity using the example with two class labels: positive and negative. We denote that there are $m$ samples of the positive class, among them $n$ samples are correctly predicted; $Acc(Y=1)=\frac{n}{m}$. We have proved that the $PCC$ loss term in equation~\ref{ap} ensures equal opportunity; two groups $a_i$ and $a_j$ have the same accuracy within subsets of samples of $Y=1$, therefore $Acc(Y=1) = Acc(Y=1|a_i) = Acc (Y=1|a_j)$, which can be rewritten as: $Acc(Y=1|a_i) = \frac{\mu n}{\mu m}, Acc(Y=1|a_i) = \frac{(1-\mu) n}{(1-\mu) m}$. 

During training, we address target class imbalance by adjusting sample probabilities to ensure an equal number of samples from each class in every training batch, so the number of samples of $Y=0$ is also denoted as $m$. The $Acc$ loss term in equation~\ref{ap} encourages $Acc(Y=1) = Acc(Y=0) = \frac{n}{m}$. Similarly, this encourages the same group accuracy for the subsets of samples of $Y=0$, i.e., $Acc(Y=0|a_i) = \frac{\phi n}{\phi m}, Acc(Y=0|a_j) = \frac{(1-\phi) n}{(1-\phi) m}$.

The overall prediction accuracy of each group is: $Acc(g_i) = \frac{\mu n + \phi n}{\mu m + \phi m} = \frac{n}{m}, Acc(g_j) = \frac{(1-\mu) n + (1-\phi) n}{(1-\mu) m + (1-\phi) m} = \frac{n}{m}$. $Acc(g_i) = Acc(g_j)$, which proves that our loss function $\mathcal{L}_{AP}$ directly optimizes for  accuracy parity.

\section{Experiment}



\begin{table*}[hbt!]
\begin{center}
  \small
  \setlength{\tabcolsep}{7.1pt}
  \scalebox{0.96}{
  \begin{tabular}{llllllll}
    \toprule
    Target & Sensitive &  \multicolumn{1}{c}{ERM} & \multicolumn{1}{c}{JTT~\cite{liu2021just}} & \multicolumn{1}{c}{BPA~\cite{seo2022unsupervised}} & \multicolumn{1}{c}{TAB~\cite{espinosa2024efficient}} & \multicolumn{1}{c}{DRO~\cite{sagawa2019distributionally} (pseudo)} &
    \multicolumn{1}{c}{Ours} \\
    \cmidrule(lr){1-2} \cmidrule(lr){3-8}
    Lipstick & Young & 0.0327 \scriptsize{(0.016)}  & 0.0483 \scriptsize{(0.0037)}  & 0.0273 \scriptsize{(0.0051)} & 0.0481 \scriptsize{(0.052)} & \underline{0.0134} \scriptsize{(0.011)} & \textbf{9.73e-3} \scriptsize{(0.0056)} \\
     Bangs & Male & 0.0960 \scriptsize{(0.0045)} & 0.0790 \scriptsize{(0.0014)}  & \underline{0.0223} \scriptsize{(0.012)}  & 0.0888 \scriptsize{(0.0097)} & 0.0275 \scriptsize{(0.0021)} & \textbf{0.0106} \scriptsize{(0.0035)} \\
    Eyeglasses & Male & \underline{0.0113} \scriptsize{(0.0025)} & 0.0130 \scriptsize{(0.0020)} & 0.0160 \scriptsize{(0.0065)} & 0.0154 \scriptsize{(0.017)} & 0.0121 \scriptsize{(0.0015)} & \textbf{9.37e-3} \scriptsize{(9.0e-4)}\\
    Blond hair & Male & 0.493 \scriptsize{(0.035)} & 0.151 \scriptsize{(0.014)}  & \underline{0.0957} \scriptsize{(0.015)}  & 0.391 \scriptsize{(0.024)} & 0.0974 \scriptsize{(0.0055)} & \textbf{0.0942} \scriptsize{(0.012)}  \\
    Wavy hair & Male & 0.385 \scriptsize{(0.021)} & 0.348 \scriptsize{(9.4e-4)} & 0.269 \scriptsize{(0.018)} & 0.289 \scriptsize{(0.063)} & \underline{0.211} \scriptsize{(0.042)} & \textbf{0.151} \scriptsize{(0.043)} \\
    Attractive & Male & 0.257 \scriptsize{(0.0072)} & 0.0670 \scriptsize{(0.025)}  & 0.101 \scriptsize{(0.026)}  & 0.128 \scriptsize{(0.038)}  & \underline{0.0530} \scriptsize{(0.017)} & \textbf{0.0420} \scriptsize{(0.013)} \\
    \bottomrule
  \end{tabular}}
  \caption{The results for the fairness notion of \textbf{equal opportunity} ($EOp$), evaluated using the recognition targets (Target) that have a naturally positive class across sensitive groups (Sensitive), on the CelebA dataset. 
The best performing method is marked in bold and the second-best method is underlined. All experiments are conducted over three runs, with average results reported to three significant figures and standard deviations to two significant figures (in parentheses). }
\label{tab:eop}
\end{center}
\end{table*}

\begin{table*}[hbt!]
\begin{center}
  \small
  \setlength{\tabcolsep}{7.1pt}
  \scalebox{0.96}{
  \begin{tabular}{llllllll}
    \toprule
    Target & Sensitive & \multicolumn{1}{c}{ERM} & \multicolumn{1}{c}{JTT~\cite{liu2021just}} & \multicolumn{1}{c}{BPA~\cite{seo2022unsupervised}} & \multicolumn{1}{c}{TAB~\cite{espinosa2024efficient}} & \multicolumn{1}{c}{DRO~\cite{sagawa2019distributionally} (pseudo)} &
    \multicolumn{1}{c}{Ours}  \\
    \cmidrule(lr){1-2} \cmidrule(lr){3-8}
     Lipstick & Young & 0.0362 \scriptsize{(0.0052)}  & 0.0539 \scriptsize{(4.6e-4)}  & 0.0322 \scriptsize{(0.0012)} & 0.0510 \scriptsize{(0.027)} & \underline{0.0305} \scriptsize{(0.020)} & \textbf{0.0256} \scriptsize{(0.0039)} \\
     Bangs & Male & 0.0685 \scriptsize{(0.0032)} &  0.0559 \scriptsize{(1.0e-3)} & 0.0430 \scriptsize{(0.0034)}  & 0.0657 \scriptsize{(0.0072)} & \underline{0.0211} \scriptsize{(0.0015)} & \textbf{0.0195} \scriptsize{(9.6e-4)} \\
     Eyeglasses & Male & \textbf{8.33e-3} \scriptsize{(0.0017)} & 0.0108 \scriptsize{(0.0014)} & 0.0206 \scriptsize{(0.0049)} & 0.0129 \scriptsize{(0.011)} & 0.0127 \scriptsize{(0.0062)} & \underline{0.0100} \scriptsize{(0.0011)} \\
     Blond hair & Male & 0.349 \scriptsize{(0.025)} & 0.133 \scriptsize{(0.0048)}  & 0.0928 \scriptsize{(0.010)}  & 0.277 \scriptsize{(0.017)}  & \underline{0.0619} \scriptsize{(0.0039)}& \textbf{0.0608} \scriptsize{(0.0059)} \\
     Wavy hair & Male & 0.275 \scriptsize{(0.015)} & 0.252 \scriptsize{(2.6e-4)} & 0.201 \scriptsize{(0.018)} & 0.208 \scriptsize{(0.045)} & \underline{0.182} \scriptsize{(0.044)} & \textbf{0.134} \scriptsize{(0.029)}\\
     Attractive & Male & 0.253 \scriptsize{(0.0095)} & \underline{0.0510} \scriptsize{(0.013)}  & 0.115 \scriptsize{(0.024)}  & 0.158 \scriptsize{(0.021)}  & 0.0605 \scriptsize{(0.019 )} & \textbf{0.0504} \scriptsize{(0.015)} \\
     \textcolor{blue!70!black}{Gender} & \textcolor{blue!70!black}{Age} & \underline{0.0609} \scriptsize{(0.0056)} & 0.0765 \scriptsize{(0.0050)} & 0.0612 \scriptsize{(0.0057)} & 0.0721 \scriptsize{(0.017)} & 0.0635 \scriptsize{(0.022)} & \textbf{0.0587} \scriptsize{(0.0032)} \\
     \textcolor{blue!70!black}{Race} & \textcolor{blue!70!black}{Gender} & 0.0172 \scriptsize{(0.0049)} & 0.0220 \scriptsize{(0.0057)} & \textbf{0.0124} \scriptsize{(0.0052)} & 0.0167 \scriptsize{(0.0089)} & 0.0145 \scriptsize{(0.0042)} & \underline{0.0127} \scriptsize{(0.0056)}\\
     \textcolor{blue!70!black}{Age} & \textcolor{blue!70!black}{Race} & 0.101 \scriptsize{(0.029)} & \underline{0.0736} \scriptsize{(0.028)} & 0.0884 \scriptsize{(0.017)} & 0.104 \scriptsize{(0.014)} & 0.0867 \scriptsize{(0.018)} & \textbf{0.0729} \scriptsize{(0.0086)}\\
     \textcolor{orange!70!black}{Species} & \textcolor{orange!70!black}{Color} & 0.0745 \scriptsize{(0.0073)}  & 0.0550 \scriptsize{(0.017)} & 0.124 \scriptsize{(0.019)} & \underline{0.0319} \scriptsize{(0.0084)} & 0.0330 \scriptsize{(0.0049)}  & \textbf{0.0297} \scriptsize{(0.0037)} \\
  \bottomrule
  \end{tabular}}
  \caption{The results for the fairness notion of \textbf{equalized odds} ($EOd$), evaluated using all classes of recognition targets (Target) across sensitive groups (Sensitive) on the CelebA dataset, \textcolor{blue!70!black}{UTKFace} dataset, and \textcolor{orange!70!black}{Dogs and Cats} dataset. }
  \vspace{-2mm}
\label{tab:eod}
\end{center}
\end{table*}

\subsection{Datasets}
\noindent We use three datasets that have been utilized for fairness and bias evaluation, strategically across multiple recognition targets and sensitive attributes. They include face image datasets CelebA~\cite{liu2015deep} and UTKFace~\cite{zhifei2017cvpr}, as in~\cite{espinosa2024efficient, hong2021unbiased, qraitem2023bias}, and Dogs and Cats dataset~\cite{Cukierski2013}  as in ~\cite{park2022fair, zhang2024distributionally} to validate the generalization of our method to general bias mitigation.

\textbf{CelebA}~\cite{liu2015deep} is a large scale dataset for facial attribute classification consisting of 200k images and 40 attributes. The demographic sensitive attributes to consider on the dataset is \textit{Male} and \textit{Young}. Following~\cite{chen2023fast, jung2022learning}, we set \textit{Blond hair} and \textit{Attractive} as the targets and \textit{Male} as the sensitive attribute. To further assess the robustness of fairness methods, we expand the target attributes to include \textit{Bangs}, \textit{Eyeglasses}, \textit{Wavy hair}, and \textit{Wearing lipstick}, which exhibit varying levels of spurious correlation to \textit{Male} and \textit{Young}~\cite{li2023partition}. The training, validation and test splits follow those provided in the original dataset. We report performance on the whole original test split.

\textbf{UTKFace}~\cite{zhifei2017cvpr} includes 23k face images in the wild and each is annotated with attributes of \textit{age} (binarized by under 35 years old and others), \textit{gender} (woman and man), and \textit{race} (White/Caucasian or not), following~\cite{park2022fair, qraitem2023bias}. We set \textit{age} as the sensitive attribute and \textit{gender} as the target as in~\cite{park2022fair, qraitem2023bias}, \textit{gender} as the sensitive attribute and \textit{race} as the target as in~\cite{park2021learning, kim2023fair}. We also evaluate the  classification for the target \textit{age} across \textit{race} sensitive attributes. We randomly select 80\% of the dataset for training, 10\% for validation, and 10\% for testing to report performance.

\textbf{Dogs and Cats}~\cite{Cukierski2013} dataset consists of 25K labeled dogs and cats images for training. The dogs and cats in the dataset are in bright or dark color, which could cause bias in model to associate the target \textit{species} with color~\cite{kim2019learning, nam2020learning, zhang2024distributionally}. We use \textit{color} as the sensitive attribute and obtain its labels from ~\cite{kim2019learning} only for evaluation. We follow the setting in ~\cite{zhang2024distributionally} to construct a biased training set; most dog images are represented in bright color and most cat images are represented in dark color, and a balanced test set.

\subsection{Baselines and Implementation}
\noindent Suitable baselines for comparison are vanilla target task learning without intervention (ERM), and fair and debiasing methods which do not use explicit supervision of sensitive attribute labels, as the setting in our work. We compare with  JTT~\cite{liu2021just} method, which upweights misclassified samples, BPA~\cite{seo2022unsupervised}  which optimizes worst-case groups recognized by latent representation clustering, and TAB~\cite{espinosa2024efficient} which balances performance of sample groups clustered by training loss history. These methods upweight or augment identified disadvantaged samples to improve model’s worst-case performance. Our method takes a different approach by directly optimizing fairness criterion within the training objective. Additionally, we compare to the fairness method DRO~\cite{sagawa2019distributionally} which requires explicit sensitive group labels but here we use \textit{discrete} VLM predicted labels (DRO (pseudo)).

We implement our method in PyTorch 2.3.1 with one NVIDIA RTX-8000 GPU. We set the regularization weight $\alpha=1$ in the main experiments and provide sensitivity analysis and a parameter-free setting in Section~\ref{sec:alpha}. We use ResNet18~\cite{he2016deep} as the classification model, trained with Adam optimizer. Data augmentations of random resized crop and horizontal flips are applied to image samples. 

\begin{table*}
\begin{center}
  \small
  \setlength{\tabcolsep}{7.2pt}
  \scalebox{0.96}{
  \begin{tabular}{lllllllll}
    \toprule
    Target & Sensitive & \multicolumn{1}{c}{ERM} & \multicolumn{1}{c}{JTT~\cite{liu2021just}} & 
    \multicolumn{1}{c} {BPA~\cite{seo2022unsupervised}} & \multicolumn{1}{c}{TAB~\cite{espinosa2024efficient}} & \multicolumn{1}{c}{DRO~\cite{sagawa2019distributionally} (pseudo)} & \multicolumn{1}{c}{Ours} \\
    \cmidrule(lr){1-2} \cmidrule(lr){3-8}
      Lipstick & Young & 0.0273 \scriptsize{(0.0027)}  &  0.0660 \scriptsize{(0.0030)} & \underline{0.0254} \scriptsize{(0.0057)} & 0.0399 \scriptsize{(0.0033)} & 0.0314 \scriptsize{(0.0037)} & \textbf{0.0242} \scriptsize{(0.0018)} \\
     Bangs & Male & \underline{0.0224} \scriptsize{(6.4e-4)} & 0.0518 \scriptsize{(6.3e-4)}  & 0.0372 \scriptsize{(0.0066)}  & 0.0308 \scriptsize{(9.5e-4)} & 0.0377 \scriptsize{(0.0012)} & \textbf{0.0182} \scriptsize{(0.0058)}  \\
     Eyeglasses & Male & \textbf{7.09e-3} \scriptsize{(4.9e-4)} & 0.0102 \scriptsize{(2.2e-3)} & 0.0128 \scriptsize{(0.0046)} & 0.0153 \scriptsize{(0.0018)} & 0.0100 \scriptsize{(2.5e-3)} & \underline{9.90e-3} \scriptsize{(0.0018)} \\
     Blond hair & Male & \underline{0.0395} \scriptsize{(0.0033)} & 0.0808 \scriptsize{(0.0034)}  & 0.0601 \scriptsize{(0.0094)}  & 0.0405 \scriptsize{(0.0011)} & 0.0441 \scriptsize{(0.0043)} & \textbf{0.0201} \scriptsize{(0.0055)} \\
     Wavy hair & Male & 0.0793 \scriptsize{(0.0058)} & 0.0675 \scriptsize{(0.0016)} & 0.127 \scriptsize{(0.036)} & 0.0925 \scriptsize{(0.0019)} & \textbf{0.0320} \scriptsize{(0.024)} & \underline{0.0503} \scriptsize{(0.011)} \\
     Attractive & Male & 0.0274 \scriptsize{(0.0024)} & 0.0228 \scriptsize{(0.0032)}  & 0.0264 \scriptsize{(0.014)}  & 0.0347 \scriptsize{(0.013)}  & \underline{0.0206} \scriptsize{(0.0056)}  & \textbf{0.0179} \scriptsize{(0.0072)}  \\
    \textcolor{blue!70!black}{Gender} & \textcolor{blue!70!black}{Age} & \underline{0.0563} \scriptsize{(0.0075)} & 0.0599 \scriptsize{(0.0091)} & 0.0572 \scriptsize{(0.0048)} & 0.0578 \scriptsize{(0.0049)} & 0.0590 \scriptsize{(0.0035)} & \textbf{0.0550} \scriptsize{(0.0025)} \\
     \textcolor{blue!70!black}{Race} & \textcolor{blue!70!black}{Gender} & \textbf{2.85e-3} \scriptsize{(6.0e-4)} & 6.82e-3 \scriptsize{(0.0060)} & 7.93e-3 \scriptsize{(0.0087)} & 5.32e-3 \scriptsize{(0.0028)} & 4.01e-3 \scriptsize{(0.0019)} & \underline{3.10e-3} \scriptsize{(0.0031)} \\
     \textcolor{blue!70!black}{Age} & \textcolor{blue!70!black}{Race} & 6.08e-3 \scriptsize{(4.6e-3)} & 4.95e-3 \scriptsize{(0.0012)} & 0.0208 \scriptsize{(0.0031)} & \underline{2.49e-3} \scriptsize{(0.0011)} & 2.88e-3 \scriptsize{(0.0056)} & \textbf{2.00e-3} \scriptsize{(0.0018)} \\
     \textcolor{orange!70!black}{Species} & \textcolor{orange!70!black}{Color} & 0.0156 \scriptsize{(0.0072)} & 6.67e-3 \scriptsize{(0.0065)} & 0.0636 \scriptsize{(0.0099)} & 7.10e-3 \scriptsize{(0.0038)} & \underline{5.92e-3} \scriptsize{(0.0079)} & \textbf{4.73e-3} \scriptsize{(0.0061)} \\
    \bottomrule
  \end{tabular}}
  \caption{The results for the fairness notion of \textbf{accuracy parity} ($AP$), evaluated by the overall accuracy for classifying targets (Target) of each sensitive attribute group (Sensitive), on the CelebA dataset, \textcolor{blue!70!black}{UTKFace} dataset, and \textcolor{orange!70!black}{Dogs and Cats} dataset. }
\label{tab:acc}
\end{center}
\end{table*}

\subsection{Main Results}
\label{sec:main}

\noindent Results on CelebA, UTKFace, and Dogs and Cats datasets are shown in Tables~\ref{tab:eop}, \ref{tab:eod}, and \ref{tab:acc}. The selected targets exhibit known biases within these datasets and are carefully chosen to represent both strong and mild bias scenarios. We evaluate fairness using equal opportunity, equalized odds, and parity accuracy (as defined in Section~\ref{sec:problem}). Because equal opportunity focuses on prediction equity for the positive ("advantaged") class, this metric is evaluated only on the CelebA dataset, where the targets naturally include a positive class. As shown in Tables~\ref{tab:eop} and \ref{tab:eod}, ERM demonstrates significant unfairness, with large disparities in predication probabilities for target classes across sensitive groups. An exception is the \textit{Eyeglasses} recognition, whose class distribution is highly imbalanced (a positive-to-negative ratio of 1:15). Here, ERM overfits to the negative class, leading to a deceptively favorable equalized odds score. Notably, our method consistently outperforms both ERM and worst-case optimization approaches across all fairness metrics. In addition, when we provide existing method with VLM-predicted group labels (DRO (pseudo)), it shows improvement over vanilla baselines but still underperforms our approach. This shows that our advantage is not merely from access to VLM information, but from our principled use of soft relevancy scores as regularization signals.

As shown in Table~\ref{tab:acc}, ERM performs well on accuracy parity and is even comparable to bias mitigation methods in some tasks. This is  because the metric can show similar overall accuracies even when groups differ in their class-specific performance. For example, one group may achieve higher accuracy on positive predictions but lower on negatives, while another shows the reverse, resulting in comparable averages despite underlying disparities. Our training objective, $\mathcal{L}_{AP}$, which is proven to optimize group-level accuracy parity, improves this metric or nearly matches the top performer. In sum, our method demonstrates robust fairness performance across recognition tasks with varying degrees of bias, datasets, and sensitive attributes.

\section{Additional Analysis}

\subsection{Robustness to VLM Errors}
\label{sec:robustness_VLM}
\begin{table}[tb!]
  \centering
  \small
  \setlength{\tabcolsep}{4pt}
  \scalebox{0.94}{
  \begin{tabular}{@{}lllll@{}}
    \toprule
    Regularization signal & $EOp$ ($\downarrow$) & $EOd$ ($\downarrow$) & $AP$ ($\downarrow$)\\
    \midrule
     ERM  & 0.493 \scriptsize(0.035) & 0.349 \scriptsize(0.025) & 0.0395 \scriptsize(0.0033)\\
     Random  & 0.459 \scriptsize(0.081) & 0.329 \scriptsize(0.054) & 0.0552 \scriptsize(0.0063)\\
     VLM + 30\% noise & 0.146 \scriptsize(0.020) & 0.116 \scriptsize(0.011) & 0.0212
     \scriptsize(0.0012) \\
     VLM (Ours) & \underline{0.0942} \scriptsize(0.012) & \textbf{0.0608} \scriptsize(0.0059) & \textbf{0.0132} \scriptsize(0.0026)\\
    Ground truth & \textbf{0.0934} \scriptsize(0.007) & \underline{0.0688} \scriptsize(0.0075) & \underline{0.0154} \scriptsize(0.0070)\\
  \bottomrule 
  \end{tabular}}
  \caption{Impact of attribute signal quality on fairness optimization for blond hair recognition on CelebA dataset. We compare: ERM (no fairness optimization), random attribute scores, VLM scores with added Gaussian noise ($\sigma=0.3$), clean VLM scores (our method), and ground truth gender attribute labels. }
  \label{tab:robustness}
\end{table}

A natural concern when using pre-trained VLMs is if our approach might inherit errors or biases in these models. It is important to clarify that our method \textbf{differs} fundamentally from using VLMs for attribute classification, since we do not use or require perfect attribute predictions. Instead, we employ VLM-generated signals as soft regularization guidance during training. Even imperfect signals that capture partial correlation with sensitive attributes can effectively guide model optimization toward fairer outcomes.

Table~\ref{tab:robustness} demonstrates this robustness empirically. On the blond hair classification task, VLM-predicted gender relevancy scores achieve substantial fairness improvements (e.g., $EOp$: 0.0942), comparable to the performance using ground truth gender labels ($EOp$: 0.0934). Critically, Importantly, our method remains effective even when VLM scores are corrupted: adding 30\% Gaussian noise to VLM scores still yields significant fairness improvements ($EOp$: 0.146) compared to the baseline ($EOp$: 0.493). In contrast, completely random attribute signals provide negligible benefit ($EOp$: 0.459), confirming that VLM predictions contain meaningful signals despite their imperfections. \textit{This robustness suggests that our regularization framework can leverage approximate attribute correlations without requiring accurate attribute prediction}.

Furthermore, our framework is agnostic to the source of attribute relevancy scores. VLMs represent one practical option when ground truth labels are unavailable, but alternative sources could be employed. The key insight is that VLM biases would only be problematic if they led to worse fairness outcomes compared to no intervention. Our results clearly demonstrate the opposite: despite potential errors, VLM-based regularization consistently improves fairness metrics across various scenarios (Tables~\ref{tab:eop}, \ref{tab:eod}, and~\ref{tab:acc}).

\subsection{Model Average Accuracy}

\begin{figure}[tb!]
  \centering
  \includegraphics[width=\linewidth]{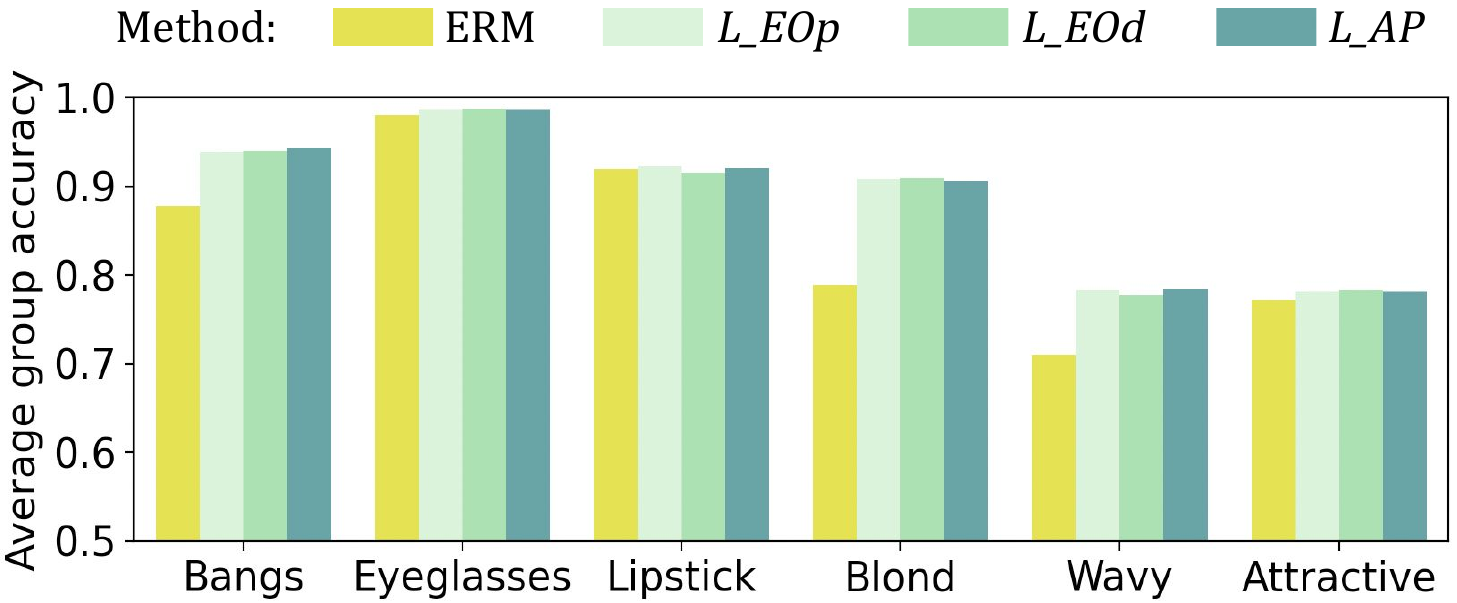}
  \caption{Comparing the average group level test accuracy of ERM method, and our proposed methods optimizing for equal opportunity $\mathcal{L}_{EOp}$, equalized odds $\mathcal{L}_{EOd}$, and accuracy parity $\mathcal{L}_{AP}$, across six attribute recognition tasks on the CelebA dataset.}
  \label{fig:wst}
\end{figure}

\noindent A concern of regularizing model training is the risk of degrading model overall quality by achieving equal predictions. We report the average group accuracy of the classification models trained with our proposed objectives. As shown in Figure~\ref{fig:wst}, though our method does not focus on improving the overall accuracy, the proposed training objectives $\mathcal{L}_{EOp}$, $\mathcal{L}_{EOd}$, and  $\mathcal{L}_{AP}$ consistently achieve higher average group accuracy than ERM across all tasks.  This demonstrates that our method enhances fairness without compromising but improving overall model quality. The improvement is partly attributed to sample reweighting to balance target classes. The strategy not only enhances overall accuracy but also supports the used fairness constraints, by ensuring that sensitive attribute-independent features are effectively learned for the underrepresented class. 

\subsection{Sensitivity to Regularization Strengths}
\label{sec:alpha}
 \begin{table}[tb!]
  \centering
  \small
  \setlength{\tabcolsep}{5pt}
  \scalebox{0.95}{
  \begin{tabular}{@{}lllll@{}}
    \toprule
    Dataset  & $\alpha$ & $EOp$ ($\downarrow$) & $EOd$ ($\downarrow$) & $AP$ ($\downarrow$)\\
    \midrule
     & 0.1 & 0.188 \scriptsize(0.048) & 0.126 \scriptsize(0.0085) & 0.0331 \scriptsize(0.0051)\\
     T: \textit{Blond} & 0.5 & 0.115 \scriptsize(0.023) & 0.0674 \scriptsize(0.0069) & 0.0193 \scriptsize(0.0025)\\
     S: \textit{Male} & 1.0 & 0.0942 \scriptsize(0.012) & \textbf{0.0608} \scriptsize(0.0059) & 0.0201 \scriptsize(0.055) \\
     & 1.5 & 0.0674 \scriptsize(0.0014) & 0.0692 \scriptsize(0.011) & 0.0164 \scriptsize(0.0068)\\
     & 2.0 & \textbf{0.0668} \scriptsize(0.022) & \underline{0.0645} \scriptsize(0.012) & \underline{0.0153} \scriptsize(0.0044)\\
     & \cellcolor{black!8}A & \cellcolor{black!8}\underline{0.0671} \scriptsize(0.020) & \cellcolor{black!8}0.0676 \scriptsize(0.019) & \cellcolor{black!8}\textbf{0.0132} \scriptsize(0.0026)\\
    \midrule
     & 0.1 & 0.0193 \scriptsize(0.0034) & 0.0301 \scriptsize(0.0076) & \textbf{0.0235} \scriptsize(0.0019) \\
    T: \textit{Lipstick} & 0.5 & \underline{8.27e-3} \scriptsize(0.010) & 0.0264 \scriptsize(0.0033) & 0.0287 \scriptsize(0.0018)\\
    S: \textit{Young} & 1.0 & 9.73e-3 \scriptsize(0.0056) & \textbf{0.0256} \scriptsize(0.0039) & \underline{0.0242} \scriptsize(0018) \\
     & 1.5 & 0.0107 \scriptsize(0.0069) & \underline{0.0262} \scriptsize(0.0056) & 0.0297 \scriptsize(0.0018)\\
     & 2.0 & 0.0116 \scriptsize(0.0029) & 0.0323 \scriptsize(0.0053) & 0.0316 \scriptsize(0.0024) \\
     & \cellcolor{black!8}A &  \cellcolor{black!8} \textbf{7.77e-3} \scriptsize(0.0049) & \cellcolor{black!8} 0.0277 \scriptsize(0.0056) &  \cellcolor{black!8}0.0257 \scriptsize(0.0035)\\
  \bottomrule 
  \end{tabular}}
  \caption{Fairness results of $EOp$, $EOd$, and $AP$ using different regularization weights ($\alpha$) in the proposed training objectives. The value of $\alpha$ can also be adaptively set without manual tuning ($\alpha$=A). Experiments are conducted on the CelebA dataset with two pairs of recognition targets (T) and sensitive attributes (S).
  }
  \label{tab:sensitivity_alpha}
\end{table}
\noindent In Table~\ref{tab:sensitivity_alpha}, we show the fairness improvements using different hyperparameter values in our method. Our approach includes a single hyperparameter to adjust, $\alpha$, which controls the weight of the fairness regularization term in the training objective (Equation~\ref{eop}, \ref{eod}, \ref{ap}). We evaluate two cases: (1) the target \textit{Blond hair} and the sensitive attribute \textit{Male} with strong spurious correlation, and (2) the target \textit{Wearing lipstick} and the sensitive attribute \textit{Young} with a weaker correlation. Results show that our method is robust to regularization strength variations: it consistently improves fairness over ERM across all $\alpha$ values with small performance shifts (Table~\ref{tab:eop}, \ref{tab:eod}, and \ref{tab:acc}). 
However, the optimal $\alpha$ might vary depending on the degree of dataset bias for recognition tasks. Specifically, overly small regularization (\textit{e.g.}, $\alpha=0.1$) may be less effective for highly biased targets, such as \textit{Blond hair}. Conversely, overly large regularization (\textit{e.g.}, $\alpha=2$ for \textit{Lipstick} recognition) may increase the risk of over-fitting to fairness constraints, potentially harming generalization and predictive accuracy. This aligns with trade-off in multi-objective optimization~\cite{sener2018multi}. An alternative to manually fine-tuning the parameter is to integrate our method with adaptive techniques, \textit{e.g.}, GradNorm~\cite{chen2018gradnorm}, which dynamically balances loss and gradient magnitudes of different tasks in the training objective. As the shaded results in Table~\ref{tab:sensitivity_alpha}, adaptively adjusting the regularization strength can achieve comparable fairness performance while making our method a hyperparameter-free pipeline.


\subsection{Ablation on Vision Language Models}

\begin{table}[]
  \centering
  \small
  \setlength{\tabcolsep}{6pt}
  \scalebox{0.94}{
  \begin{tabular}{@{}llll@{}}
    \toprule
    VLM & Target / Sensitive  & $EOd$ ($\downarrow$) & $AP$ ($\downarrow$)  \\
    \midrule
     & \textit{Blond / Male}  & 0.0853 \scriptsize(0.013)  & 0.0512 \scriptsize(0.0027)\\
    BLIP~\cite{li2023blip} & \textit{Gender / Age} & 0.0603 \scriptsize(0.0018) & 0.0524 \scriptsize(0.0022) \\
     & \textit{Species / Color}  & 0.0341 \scriptsize(0.0020) & 6.25e-3 \scriptsize(0.0055) \\
    \midrule
    & \textit{Blond / Male} & 0.0700 \scriptsize(0.020) & 0.0108 \scriptsize(0.0079) \\
    ALIGN~\cite{jia2021scaling} & \textit{Gender / Age} & 0.0547 \scriptsize(0.010) & 0.0543 \scriptsize(0.0082) \\
    & \textit{Species / Color} & 0.0340 \scriptsize(0.0073) & 8.07e-3 \scriptsize(0.0068)\\
  \bottomrule
  \end{tabular}}
  \caption{The vision-language model, BLIP and ALIGN, are used to implement our fairness objectives. Results of $EOd$ and $AP$ are reported for the three pairs of target and sensitive attributes on the CelebA, UTKFace, and Dogs and Cats dataset respectively ($EOp$ is not available for most targets so it is not included here).    
  }
  \label{tab:BLIP}
\end{table}

\noindent We replace the CLIP model used in the main experiments with other state-of-the-art VLMs, BLIP~\cite{li2023blip} that enhances fine-grained understanding on image-text relationship, and ALIGN~\cite{jia2021scaling} that was pre-trained on larger-scale data, to provide alignment scores between image samples and text descriptions for sensitive attributes. Shown in Table~\ref{tab:BLIP}, fairness regularization based on BLIP and ALIGN predictions remains effective in improving the metrics, comparable to that on CLIP (see Section~\ref{sec:main}), though the results for \textit{Blond} recognition based on BLIP are slightly worse than the other two models. Our method is agnostic to the vision-language model used if the model can reasonably approximate how an attribute is perceived in an image (Assumption 1).

\subsection{Generalization to Multi-Category Attribute}
\label{sec:multipleG}

\begin{table}[tb!]
  \centering
  \small
  \setlength{\tabcolsep}{10pt}
  \scalebox{0.93}{
  \begin{tabular}{lll}
    \toprule
    Method &  $EOd$ ($\downarrow$) & $AP$ ($\downarrow$)\\
    \midrule
     ERM &   0.296(0.038) & 0.183 (0.034)\\
      JTT &  \underline{0.150} (0.020) & 0.173 (0.045) \\
      BPA &  0.176 (0.033) & 0.148 (0.051)\\
      TAB & 0.230 (0.0057) & \underline{0.138} (0.027)\\
       Ours &  \textbf{0.130} (0.0040) & \textbf{0.0988} (0.019)\\
    \bottomrule
  \end{tabular}}
  \caption{Comparison of fairness metrics between the baseline and our proposed methods across multiple \textbf{\textit{race}} groups: \textit{White}, \textit{Black}, \textit{Asian}, and \textit{Indian} for \textbf{\textit{age}} recognition on the UTKFace dataset.
  }
  \label{tab:multipleS}
\end{table}

\noindent To evaluate our method’s ability to improve fairness across non-binary groups, we use the \textit{race} attribute from the UTKFace dataset to be the sensitive attribute, because it includes annotations for more than two categories. The recognition target is \textit{Age} as in main experiments. Table~\ref{tab:multipleS} shows that in the special case where our training objectives regularize the correlation between the predicted class probability and attribute relevancy in all attribute categories, our method still effectively promotes parity across multiple sensitive groups, outperforming baseline bias mitigation methods.

\section{Conclusion}
\noindent In this work, we propose a paradigm shift from proxy-based to direct optimization of fairness metrics in image classification. Our training objectives mathematically connect to specific fairness criteria, enabling practitioners to directly optimize the metric most relevant to their application. We combine target task loss with regularization terms that encourage independence between model predictions and VLM-analyzed attribute relevancy scores, providing flexible fairness optimization without group annotations. Experimental results on three image datasets validate the effectiveness of direct metric optimization, showing consistent improvements on targeted fairness metrics, while maintaining robustness to VLM imperfections.

\noindent \textbf{Limitations.} The current framework focuses on parity-based fairness metrics (equal opportunity, equalized odds, accuracy parity). Other notions like individual fairness, counterfactual fairness, or calibration-based metrics would require different mathematical formulations and may not fit as naturally into this correlation-based regularization framework. Additionally, while our method is robust to VLM errors (Section~\ref{sec:robustness_VLM}), the method performs best when VLMs can leverage their pre-trained knowledge about visual concepts. For specialized domains that differ significantly from typical VLM training data, such as medical imaging, the attribute relevancy scores may be less informative. Our framework's flexibility allows for incorporating domain-specific models or alternative attribute signals when needed, making it adaptable to diverse application contexts.

{\small
\bibliographystyle{ieee_fullname}
\bibliography{egbib}

\begin{thebibliography}{10}\itemsep=-1pt

\bibitem{abdelfattah2023cdul}
Rabab Abdelfattah, Qing Guo, Xiaoguang Li, Xiaofeng Wang, and Song Wang.
\newblock Cdul: Clip-driven unsupervised learning for multi-label image classification.
\newblock In {\em Proceedings of the IEEE/CVF International Conference on Computer Vision}, pages 1348--1357, 2023.

\bibitem{agarwal2022does}
Sharat Agarwal, Sumanyu Muku, Saket Anand, and Chetan Arora.
\newblock Does data repair lead to fair models? curating contextually fair data to reduce model bias.
\newblock In {\em Proceedings of the IEEE/CVF Winter Conference on Applications of Computer Vision}, pages 3298--3307, 2022.

\bibitem{Ahn2023mitigating}
Sumyeong Ahn, Seongyoon Kim, and Se-young Yun.
\newblock Mitigating dataset bias by using per-sample gradient.
\newblock In {\em Eleventh International Conference on Learning Representations}. ICLR, 2023.

\bibitem{berk2021fairness}
Richard Berk, Hoda Heidari, Shahin Jabbari, Michael Kearns, and Aaron Roth.
\newblock Fairness in criminal justice risk assessments: The state of the art.
\newblock {\em Sociological Methods \& Research}, 50(1):3--44, 2021.

\bibitem{bommasani2021opportunities}
Rishi Bommasani, Drew~A Hudson, Ehsan Adeli, Russ Altman, Simran Arora, Sydney von Arx, Michael~S Bernstein, Jeannette Bohg, Antoine Bosselut, Emma Brunskill, et~al.
\newblock On the opportunities and risks of foundation models.
\newblock {\em arXiv preprint arXiv:2108.07258}, 2021.

\bibitem{10.1145/3616865}
Simon Caton and Christian Haas.
\newblock Fairness in machine learning: A survey.
\newblock {\em ACM Comput. Surv.}, 56(7), Apr. 2024.

\bibitem{chai2022fairness}
Junyi Chai, Taeuk Jang, and Xiaoqian Wang.
\newblock Fairness without demographics through knowledge distillation.
\newblock {\em Advances in Neural Information Processing Systems}, 35:19152--19164, 2022.

\bibitem{chakraborty2024exmap}
Rwiddhi Chakraborty, Adrian Sletten, and Michael~C Kampffmeyer.
\newblock Exmap: Leveraging explainability heatmaps for unsupervised group robustness to spurious correlations.
\newblock In {\em Proceedings of the IEEE/CVF Conference on Computer Vision and Pattern Recognition}, pages 12017--12026, 2024.

\bibitem{chen2023fast}
Ruizhe Chen, Jianfei Yang, Huimin Xiong, Jianhong Bai, Tianxiang Hu, Jin Hao, Yang Feng, Joey~Tianyi Zhou, Jian Wu, and Zuozhu Liu.
\newblock Fast model debias with machine unlearning.
\newblock {\em Advances in Neural Information Processing Systems}, 36:14516--14539, 2023.

\bibitem{chen2018gradnorm}
Zhao Chen, Vijay Badrinarayanan, Chen-Yu Lee, and Andrew Rabinovich.
\newblock Gradnorm: Gradient normalization for adaptive loss balancing in deep multitask networks.
\newblock In {\em International conference on machine learning}, pages 794--803. PMLR, 2018.

\bibitem{cohen2009pearson}
Israel Cohen, Yiteng Huang, Jingdong Chen, Jacob Benesty, Jacob Benesty, Jingdong Chen, Yiteng Huang, and Israel Cohen.
\newblock Pearson correlation coefficient.
\newblock {\em Noise reduction in speech processing}, pages 1--4, 2009.

\bibitem{Cukierski2013}
Will Cukierski.
\newblock Dogs vs. cats, 2013.
\newblock Kaggle competition.

\bibitem{d2024improving}
Moreno D'Inc{\`a}, Christos Tzelepis, Ioannis Patras, and Nicu Sebe.
\newblock Improving fairness using vision-language driven image augmentation.
\newblock In {\em Proceedings of the IEEE/CVF Winter Conference on Applications of Computer Vision}, pages 4695--4704, 2024.

\bibitem{dooley2022robustness}
Samuel Dooley, George~Z Wei, Tom Goldstein, and John Dickerson.
\newblock Robustness disparities in face detection.
\newblock {\em Advances in Neural Information Processing Systems}, 35:38245--38259, 2022.

\bibitem{dunlap2023using}
Lisa Dunlap, Clara Mohri, Devin Guillory, Han Zhang, Trevor Darrell, Joseph~E Gonzalez, Aditi Raghunathan, and Anja Rohrbach.
\newblock Using language to extend to unseen domains.
\newblock In {\em Eleventh International Conference on Learning Representations}. ICLR, 2023.

\bibitem{espinosa2024efficient}
Mateo Espinosa~Zarlenga, Swami Sankaranarayanan, Jerone~TA Andrews, Zohreh Shams, Mateja Jamnik, and Alice Xiang.
\newblock Efficient bias mitigation without privileged information.
\newblock In {\em European Conference on Computer Vision}, pages 148--166. Springer, 2024.

\bibitem{garg2020fairness}
Pratyush Garg, John Villasenor, and Virginia Foggo.
\newblock Fairness metrics: A comparative analysis.
\newblock In {\em 2020 IEEE international conference on big data (Big Data)}, pages 3662--3666. IEEE, 2020.

\bibitem{gong2021mitigating}
Sixue Gong, Xiaoming Liu, and Anil~K Jain.
\newblock Mitigating face recognition bias via group adaptive classifier.
\newblock In {\em Proceedings of the IEEE/CVF conference on computer vision and pattern recognition}, pages 3414--3424, 2021.

\bibitem{goyal2022fairness}
Priya Goyal, Adriana~Romero Soriano, Caner Hazirbas, Levent Sagun, and Nicolas Usunier.
\newblock Fairness indicators for systematic assessments of visual feature extractors.
\newblock In {\em Proceedings of the 2022 ACM Conference on Fairness, Accountability, and Transparency}, pages 70--88, 2022.

\bibitem{hardt2016equality}
Moritz Hardt, Eric Price, and Nati Srebro.
\newblock Equality of opportunity in supervised learning.
\newblock {\em Advances in neural information processing systems}, 29, 2016.

\bibitem{he2016deep}
Kaiming He, Xiangyu Zhang, Shaoqing Ren, and Jian Sun.
\newblock Deep residual learning for image recognition.
\newblock In {\em Proceedings of the IEEE conference on computer vision and pattern recognition}, pages 770--778, 2016.

\bibitem{he2023clip}
Wenbin He, Suphanut Jamonnak, Liang Gou, and Liu Ren.
\newblock Clip-s4: Language-guided self-supervised semantic segmentation.
\newblock In {\em Proceedings of the IEEE/CVF Conference on Computer Vision and Pattern Recognition}, pages 11207--11216, 2023.

\bibitem{hong2021unbiased}
Youngkyu Hong and Eunho Yang.
\newblock Unbiased classification through bias-contrastive and bias-balanced learning.
\newblock {\em Advances in Neural Information Processing Systems}, 34:26449--26461, 2021.

\bibitem{jia2021scaling}
Chao Jia, Yinfei Yang, Ye Xia, Yi-Ting Chen, Zarana Parekh, Hieu Pham, Quoc Le, Yun-Hsuan Sung, Zhen Li, and Tom Duerig.
\newblock Scaling up visual and vision-language representation learning with noisy text supervision.
\newblock In {\em International conference on machine learning}, pages 4904--4916. PMLR, 2021.

\bibitem{jung2022learning}
Sangwon Jung, Sanghyuk Chun, and Taesup Moon.
\newblock Learning fair classifiers with partially annotated group labels.
\newblock In {\em Proceedings of the IEEE/CVF conference on computer vision and pattern recognition}, pages 10348--10357, 2022.

\bibitem{jung2023re}
Sangwon Jung, Taeeon Park, Sanghyuk Chun, and Taesup Moon.
\newblock Re-weighting based group fairness regularization via classwise robust optimization.
\newblock In {\em Twelveth International Conference on Learning Representations}. ICLR, 2023.

\bibitem{kim2019learning}
Byungju Kim, Hyunwoo Kim, Kyungsu Kim, Sungjin Kim, and Junmo Kim.
\newblock Learning not to learn: Training deep neural networks with biased data.
\newblock In {\em Proceedings of the IEEE/CVF conference on computer vision and pattern recognition}, pages 9012--9020, 2019.

\bibitem{kim2023fair}
Dohyung Kim, Sungho Park, Sunhee Hwang, and Hyeran Byun.
\newblock Fair classification by loss balancing via fairness-aware batch sampling.
\newblock {\em Neurocomputing}, 518:231--241, 2023.

\bibitem{kim2023bias}
Younghyun Kim, Sangwoo Mo, Minkyu Kim, Kyungmin Lee, Jaeho Lee, and Jinwoo Shin.
\newblock Bias-to-text: Debiasing unknown visual biases through language interpretation.
\newblock {\em arXiv preprint arXiv:2301.11104}, 2, 2023.

\bibitem{kim2024discovering}
Younghyun Kim, Sangwoo Mo, Minkyu Kim, Kyungmin Lee, Jaeho Lee, and Jinwoo Shin.
\newblock Discovering and mitigating visual biases through keyword explanation.
\newblock In {\em Proceedings of the IEEE/CVF Conference on Computer Vision and Pattern Recognition}, pages 11082--11092, 2024.

\bibitem{krishnakumar2021udis}
Arvindkumar Krishnakumar, Viraj Prabhu, Sruthi Sudhakar, and Judy Hoffman.
\newblock Udis: Unsupervised discovery of bias in deep visual recognition models.
\newblock In {\em British Machine Vision Conference (BMVC)}, volume~1, page~3, 2021.

\bibitem{lahoti2020fairness}
Preethi Lahoti, Alex Beutel, Jilin Chen, Kang Lee, Flavien Prost, Nithum Thain, Xuezhi Wang, and Ed Chi.
\newblock Fairness without demographics through adversarially reweighted learning.
\newblock {\em Advances in neural information processing systems}, 33:728--740, 2020.

\bibitem{li2022cat}
Jiazhi Li and Wael Abd-Almageed.
\newblock Cat: Controllable attribute translation for fair facial attribute classification.
\newblock In {\em European Conference on Computer Vision}, pages 363--381. Springer, 2022.

\bibitem{li2023blip}
Junnan Li, Dongxu Li, Silvio Savarese, and Steven Hoi.
\newblock Blip-2: Bootstrapping language-image pre-training with frozen image encoders and large language models.
\newblock {\em arXiv preprint arXiv:2301.12597}, 2023.

\bibitem{li2023partition}
Jiaxuan Li, Duc~Minh Vo, and Hideki Nakayama.
\newblock Partition-and-debias: Agnostic biases mitigation via a mixture of biases-specific experts.
\newblock In {\em Proceedings of the IEEE/CVF International Conference on Computer Vision}, pages 4924--4934, 2023.

\bibitem{li2019repair}
Yi Li and Nuno Vasconcelos.
\newblock Repair: Removing representation bias by dataset resampling.
\newblock In {\em Proceedings of the IEEE/CVF conference on computer vision and pattern recognition}, pages 9572--9581, 2019.

\bibitem{li2022discover}
Zhiheng Li, Anthony Hoogs, and Chenliang Xu.
\newblock Discover and mitigate unknown biases with debiasing alternate networks.
\newblock In {\em European Conference on Computer Vision}, pages 270--288. Springer, 2022.

\bibitem{lingenfelter2022quantitative}
Bryson Lingenfelter, Sara~R Davis, and Emily~M Hand.
\newblock A quantitative analysis of labeling issues in the celeba dataset.
\newblock In {\em International Symposium on Visual Computing}, pages 129--141. Springer, 2022.

\bibitem{liu2021just}
Evan~Z Liu, Behzad Haghgoo, Annie~S Chen, Aditi Raghunathan, Pang~Wei Koh, Shiori Sagawa, Percy Liang, and Chelsea Finn.
\newblock Just train twice: Improving group robustness without training group information.
\newblock In {\em International Conference on Machine Learning}, pages 6781--6792. PMLR, 2021.

\bibitem{liu2015deep}
Ziwei Liu, Ping Luo, Xiaogang Wang, and Xiaoou Tang.
\newblock Deep learning face attributes in the wild.
\newblock In {\em Proceedings of the IEEE international conference on computer vision}, pages 3730--3738, 2015.

\bibitem{liu2021image}
Zheyuan Liu, Cristian Rodriguez-Opazo, Damien Teney, and Stephen Gould.
\newblock Image retrieval on real-life images with pre-trained vision-and-language models.
\newblock In {\em Proceedings of the IEEE/CVF International Conference on Computer Vision}, pages 2125--2134, 2021.

\bibitem{lou2022tecm}
Xudong Lou, Yiguang Liu, and Xuwei Li.
\newblock Tecm-clip: Text-based controllable multi-attribute face image manipulation.
\newblock In {\em Proceedings of the Asian Conference on Computer Vision}, pages 1942--1958, 2022.

\bibitem{mashiat2022trade}
Tasfia Mashiat, Xavier Gitiaux, Huzefa Rangwala, Patrick Fowler, and Sanmay Das.
\newblock Trade-offs between group fairness metrics in societal resource allocation.
\newblock In {\em Proceedings of the 2022 ACM Conference on Fairness, Accountability, and Transparency}, pages 1095--1105, 2022.

\bibitem{nam2020learning}
Junhyun Nam, Hyuntak Cha, Sungsoo Ahn, Jaeho Lee, and Jinwoo Shin.
\newblock Learning from failure: De-biasing classifier from biased classifier.
\newblock {\em Advances in Neural Information Processing Systems}, 33:20673--20684, 2020.

\bibitem{nanda2021fairness}
Vedant Nanda, Samuel Dooley, Sahil Singla, Soheil Feizi, and John~P Dickerson.
\newblock Fairness through robustness: Investigating robustness disparity in deep learning.
\newblock In {\em Proceedings of the 2021 ACM Conference on Fairness, Accountability, and Transparency}, pages 466--477, 2021.

\bibitem{park2021learning}
Sungho Park, Sunhee Hwang, Dohyung Kim, and Hyeran Byun.
\newblock Learning disentangled representation for fair facial attribute classification via fairness-aware information alignment.
\newblock In {\em Proceedings of the AAAI Conference on Artificial Intelligence}, volume~35, pages 2403--2411, 2021.

\bibitem{park2022fair}
Sungho Park, Jewook Lee, Pilhyeon Lee, Sunhee Hwang, Dohyung Kim, and Hyeran Byun.
\newblock Fair contrastive learning for facial attribute classification.
\newblock In {\em Proceedings of the IEEE/CVF Conference on Computer Vision and Pattern Recognition}, pages 10389--10398, 2022.

\bibitem{petryk2022guiding}
Suzanne Petryk, Lisa Dunlap, Keyan Nasseri, Joseph Gonzalez, Trevor Darrell, and Anna Rohrbach.
\newblock On guiding visual attention with language specification.
\newblock In {\em Proceedings of the IEEE/CVF Conference on Computer Vision and Pattern Recognition}, pages 18092--18102, 2022.

\bibitem{qraitem2023bias}
Maan Qraitem, Kate Saenko, and Bryan~A Plummer.
\newblock Bias mimicking: A simple sampling approach for bias mitigation.
\newblock In {\em Proceedings of the IEEE/CVF Conference on Computer Vision and Pattern Recognition}, pages 20311--20320, 2023.

\bibitem{radford2021learning}
Alec Radford, Jong~Wook Kim, Chris Hallacy, Aditya Ramesh, Gabriel Goh, Sandhini Agarwal, Girish Sastry, Amanda Askell, Pamela Mishkin, Jack Clark, et~al.
\newblock Learning transferable visual models from natural language supervision.
\newblock In {\em International conference on machine learning}, pages 8748--8763. PMLR, 2021.

\bibitem{ramaswamy2021fair}
Vikram~V Ramaswamy, Sunnie~SY Kim, and Olga Russakovsky.
\newblock Fair attribute classification through latent space de-biasing.
\newblock In {\em Proceedings of the IEEE/CVF conference on computer vision and pattern recognition}, pages 9301--9310, 2021.

\bibitem{romano2020achieving}
Yaniv Romano, Stephen Bates, and Emmanuel Candes.
\newblock Achieving equalized odds by resampling sensitive attributes.
\newblock {\em Advances in neural information processing systems}, 33:361--371, 2020.

\bibitem{sagawa2019distributionally}
Shiori Sagawa, Pang~Wei Koh, Tatsunori~B Hashimoto, and Percy Liang.
\newblock Distributionally robust neural networks for group shifts: On the importance of regularization for worst-case generalization.
\newblock {\em arXiv preprint arXiv:1911.08731}, 2019.

\bibitem{sener2018multi}
Ozan Sener and Vladlen Koltun.
\newblock Multi-task learning as multi-objective optimization.
\newblock {\em Advances in neural information processing systems}, 31, 2018.

\bibitem{seo2022unsupervised}
Seonguk Seo, Joon-Young Lee, and Bohyung Han.
\newblock Unsupervised learning of debiased representations with pseudo-attributes.
\newblock In {\em Proceedings of the IEEE/CVF Conference on Computer Vision and Pattern Recognition}, pages 16742--16751, 2022.

\bibitem{shen2022optimising}
Aili Shen, Xudong Han, Trevor Cohn, Timothy Baldwin, and Lea Frermann.
\newblock Optimising equal opportunity fairness in model training.
\newblock {\em arXiv preprint arXiv:2205.02393}, 2022.

\bibitem{shen2023clip}
Shuai Shen, Wanhua Li, Xiaobing Wang, Dafeng Zhang, Zhezhu Jin, Jie Zhou, and Jiwen Lu.
\newblock Clip-cluster: Clip-guided attribute hallucination for face clustering.
\newblock In {\em Proceedings of the IEEE/CVF International Conference on Computer Vision}, pages 20786--20795, 2023.

\bibitem{torralba2011unbiased}
Antonio Torralba and Alexei~A Efros.
\newblock Unbiased look at dataset bias.
\newblock In {\em CVPR 2011}, pages 1521--1528. IEEE, 2011.

\bibitem{wang2023exploring}
Jianyi Wang, Kelvin~CK Chan, and Chen~Change Loy.
\newblock Exploring clip for assessing the look and feel of images.
\newblock In {\em Proceedings of the AAAI Conference on Artificial Intelligence}, volume~37, pages 2555--2563, 2023.

\bibitem{wang2022cris}
Zhaoqing Wang, Yu Lu, Qiang Li, Xunqiang Tao, Yandong Guo, Mingming Gong, and Tongliang Liu.
\newblock Cris: Clip-driven referring image segmentation.
\newblock In {\em Proceedings of the IEEE/CVF conference on computer vision and pattern recognition}, pages 11686--11695, 2022.

\bibitem{wang2020towards}
Zeyu Wang, Klint Qinami, Ioannis~Christos Karakozis, Kyle Genova, Prem Nair, Kenji Hata, and Olga Russakovsky.
\newblock Towards fairness in visual recognition: Effective strategies for bias mitigation.
\newblock In {\em Proceedings of the IEEE/CVF conference on computer vision and pattern recognition}, pages 8919--8928, 2020.

\bibitem{wu2023discover}
Shirley Wu, Mert Yuksekgonul, Linjun Zhang, and James Zou.
\newblock Discover and cure: Concept-aware mitigation of spurious correlation.
\newblock In {\em International Conference on Machine Learning}, pages 37765--37786. PMLR, 2023.

\bibitem{yan2020fair}
Shen Yan, Hsien-te Kao, and Emilio Ferrara.
\newblock Fair class balancing: Enhancing model fairness without observing sensitive attributes.
\newblock In {\em Proceedings of the 29th ACM International Conference on Information \& Knowledge Management}, pages 1715--1724, 2020.

\bibitem{yenamandra2023facts}
Sriram Yenamandra, Pratik Ramesh, Viraj Prabhu, and Judy Hoffman.
\newblock Facts: First amplify correlations and then slice to discover bias.
\newblock In {\em Proceedings of the IEEE/CVF International Conference on Computer Vision}, pages 4794--4804, 2023.

\bibitem{zhang2024distributionally}
Fengda Zhang, Qianpei He, Kun Kuang, Jiashuo Liu, Long Chen, Chao Wu, Jun Xiao, and Hanwang Zhang.
\newblock Distributionally generative augmentation for fair facial attribute classification.
\newblock In {\em Proceedings of the IEEE/CVF Conference on Computer Vision and Pattern Recognition}, pages 22797--22808, 2024.

\bibitem{zhang2024common}
Miao Zhang, Ben Colman, Ali Shahriyari, Gaurav Bharaj, et~al.
\newblock Common-sense bias discovery and mitigation for classification tasks.
\newblock {\em arXiv preprint arXiv:2401.13213}, 2024.

\bibitem{zhang2022tip}
Renrui Zhang, Wei Zhang, Rongyao Fang, Peng Gao, Kunchang Li, Jifeng Dai, Yu Qiao, and Hongsheng Li.
\newblock Tip-adapter: Training-free adaption of clip for few-shot classification.
\newblock In {\em European Conference on Computer Vision}, pages 493--510. Springer, 2022.

\bibitem{zhifei2017cvpr}
Song~Yang Zhang, Zhifei and Hairong Qi.
\newblock Age progression/regression by conditional adversarial autoencoder.
\newblock In {\em IEEE Conference on Computer Vision and Pattern Recognition (CVPR)}. IEEE, 2017.

\bibitem{zhang2023drml}
Yuhui Zhang, Jeff~Z HaoChen, Shih-Cheng Huang, Kuan-Chieh Wang, James Zhou, and Serena Yeung.
\newblock Drml: Diagnosing and rectifying vision models using language.
\newblock In {\em Eleventh International Conference on Learning Representations}. ICLR, 2022.

\bibitem{zhang2024discover}
Zeliang Zhang, Mingqian Feng, Zhiheng Li, and Chenliang Xu.
\newblock Discover and mitigate multiple biased subgroups in image classifiers.
\newblock In {\em Proceedings of the IEEE/CVF Conference on Computer Vision and Pattern Recognition}, pages 10906--10915, 2024.

\bibitem{zhong2022regionclip}
Yiwu Zhong, Jianwei Yang, Pengchuan Zhang, Chunyuan Li, Noel Codella, Liunian~Harold Li, Luowei Zhou, Xiyang Dai, Lu Yuan, Yin Li, et~al.
\newblock Regionclip: Region-based language-image pretraining.
\newblock In {\em Proceedings of the IEEE/CVF Conference on Computer Vision and Pattern Recognition}, pages 16793--16803, 2022.

\end{thebibliography}
}

\end{document}